\documentclass[letterpaper, 10 pt, conference]{ieeeconf}  

\IEEEoverridecommandlockouts                              

\usepackage{multicol}
\usepackage{graphicx}
\usepackage{amsmath}
\usepackage{bbm}
\usepackage{dblfloatfix}
\usepackage[ruled]{algorithm2e}

\usepackage[zerostyle=c,scaled=.94]{newtxtt}
\SetCommentSty{mycommfont}
\SetFuncSty{myprocname}

\usepackage{mathrsfs}
\usepackage{bm}
\usepackage[makeroom]{cancel}
\usepackage{amssymb}
\usepackage{upgreek}
\usepackage{multirow}
\usepackage{amsfonts}
\usepackage{here}
\usepackage[normalem]{ulem}
\usepackage{mdframed}

\newcommand{\bb}[1]{\textbf{#1}}

\usepackage{booktabs}
\usepackage{color}
\usepackage{subcaption}
\usepackage[font=small, skip=5pt]{caption}
\usepackage{listings}
\usepackage{hyperref}
\hypersetup{
    colorlinks,
    linkcolor={red!50!black},
    citecolor={blue!80!black},
    urlcolor={blue!80!black}
}
\usepackage{wrapfig}
\usepackage{pdflscape}
\usepackage{setspace}
\makeatletter
\let\NAT@parse\undefined
\makeatother
\usepackage[numbers]{natbib}

\usepackage{cleveref}
\usepackage{tikz}
\DeclareMathAlphabet{\mathcal}{OMS}{cmsy}{m}{n}
\DeclareMathAlphabet\mathbfcal{OMS}{cmsy}{b}{n}

\newcommand{\mc}[1]{\mathcal{#1}}

\DeclareMathOperator*{\argmax}{argmax}
\DeclareMathOperator*{\argmin}{argmin}

\newcommand{\E}{\mathbb{E}}

\graphicspath{{figs/}}

\newcommand{\norm}[1]{\left\lVert#1\right\rVert}
\newcommand{\eg}{\textit{e.g.}}
\newcommand{\ie}{\textit{i.e.}}

\usepackage{xcolor,colortbl}
\definecolor{Gray}{gray}{0.85}
\definecolor{LightCyan}{rgb}{0.88,1,1}
\newcolumntype{a}{>{\columncolor{Gray}}c}
\newcolumntype{b}{>{\columncolor{white}}c}

\definecolor{purple}{rgb}{0.37, 0.18, 0.67}
\definecolor{darkgreen}{rgb}{0.08, 0.55, 0.08}
\definecolor{blue1}{rgb}{0.2, 0.2, 0.6}
\definecolor{ltgreen}{rgb}{0.5, 0.8, 0.5}
\definecolor{green1}{rgb}{0.2, 0.6, 0.2}
\definecolor{green2}{rgb}{0.1, 0.4, 0.1}
\definecolor{dkgreen}{rgb}{0,0.7,0}
\definecolor{dkgreen2}{rgb}{0,0.55,0}
\definecolor{dkgreen3}{rgb}{0,0.4,0}
\definecolor{gray}{rgb}{0.5,0.5,0.5}
\definecolor{mauve}{rgb}{0.58,0,0.82}
\definecolor{magenta}{rgb}{0.82,0,0.82}

\newcommand{\omark}[2][]{{\color{dkgreen3}
    \ifthenelse{\equal{#1}{}}{}{\textsf{[#1]}:}#2}}

\definecolor{purple}{rgb}{0.858, 0.08, 0.85}
\definecolor{darkgreen}{rgb}{0.08, 0.55, 0.08}
\definecolor{blue1}{rgb}{0.2, 0.2, 0.6}
\definecolor{green1}{rgb}{0.2, 0.6, 0.2}
\definecolor{green2}{rgb}{0.1, 0.4, 0.1}

\newcommand{\mindeg}{\zeta_{\text{min}}}
\newcommand{\maxdeg}{\zeta_{\text{max}}}
\newcommand{\cosstate}{\xtarget,\srobot}
\newcommand{\fullstate}{\substack{x_1,\cdots,x_n,\\\xtarget,\srobot}}
\newcommand{\SFPOMDP}{\mc{S}_{\text{F}}}
\newcommand{\PrFPOMDP}{\Pr_{\text{F-POMDP}}}
\newcommand{\PrCOS}{\Pr_{\text{COS-POMDP}}}

\newcommand{\target}{\text{target}}

\newcommand{\zrobot}{z_{\text{robot}}}
\newcommand{\zobjects}{z_{\text{objects}}}
\newcommand{\srobot}{s_{\text{robot}}}
\newcommand{\srobotinit}{s_{\text{robot}}^{\text{init}}}
\newcommand{\xtarget}{x_{\text{target}}}
\newcommand{\btarget}{b_{\text{target}}}
\newcommand{\brobot}{b_{\text{robot}}}
\newcommand{\Srobot}{\mc{S}_{\text{robot}}}

\newcommand{\Dtarget}{D_{\text{target}}}
\newcommand{\ztarget}{z_{\text{target}}}

\newcommand{\Orobot}{O_{\text{robot}}}

\newcommand{\DECLARE}{\texttt{Done}}
\newcommand{\indicator}{\mathbbm{1}}

\newcommand{\rmax}{R_{\text{max}}}
\newcommand{\rmin}{R_{\text{min}}}

\newcommand{\dsep}{d_{\text{sep}}}

\title{\LARGE \bf
Towards Optimal Correlational Object Search}

\author{Author
  \thanks{Institution}
}

\author{Kaiyu Zheng$^{\dagger}$, Rohan Chitnis$^{*}$, Yoonchang Sung$^{*}$,
  George Konidaris$^{\dagger}$, Stefanie Tellex$^{\dagger}$
  \thanks{$^{\dagger}$Brown University, Providence, RI. $^{*}$MIT CSAIL, Cambridge, MA.}
  \thanks{Email correspondence: \{\texttt{kzheng10@cs.brown.edu}\}}%
}

\begin{document}

\maketitle
\thispagestyle{empty}
\pagestyle{empty}
\begin{abstract}
  In realistic applications of object search, robots will need to locate target
  objects in complex environments while coping with unreliable sensors,
  especially for small or hard-to-detect objects. In such settings,
  \emph{correlational information} can be valuable for planning efficiently.
  Previous approaches that consider correlational information typically resort
  to ad-hoc, greedy search strategies. We introduce the Correlational Object Search
  POMDP (COS-POMDP), which models correlations while preserving optimal solutions
  with a reduced state space. We propose a hierarchical planning algorithm to
  scale up COS-POMDPs for practical domains.  Our evaluation, conducted with the AI2-THOR
  household simulator and the YOLOv5 object detector, shows that our method finds objects more successfully and efficiently compared to baselines, particularly for hard-to-detect objects such as srub brush and remote control.
\end{abstract}

\IEEEpeerreviewmaketitle


\section{Introduction}
\begin{figure}[t]
  \centering
  \noindent
    \includegraphics[width=\columnwidth]{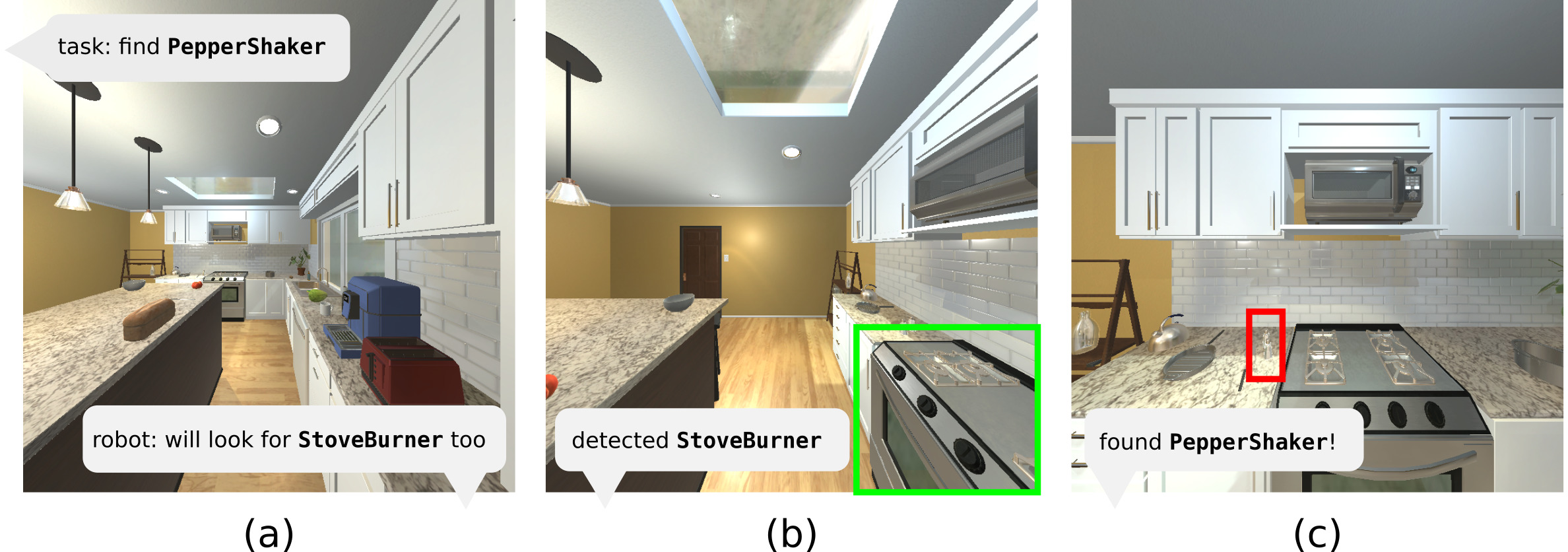}
    \caption{We study the problem of object search using correlational information about spatial relations between objects. This example illustrates a desirable search behavior in an AI2-THOR scene, where the robot leverages the detection of a large \texttt{StoveBurner} to more efficiently find a small, hard-to-detect \texttt{PepperShaker}.}
    \label{fig:teaser}
    \vspace{-0.2in}
\end{figure}

Object search is a fundamental capability for robots  in many applications
including domestic services \cite{sprute2017ambient,zeng2020semantic}, search
and rescue \cite{eismann2009automated,sun2016camera}, and elderly care
\cite{idrees2020robomem,loghmani2018towards}. In realistic settings, the object being searched for (e.g. pepper shaker) will often be small, outside the current field of view, and hard to detect. In such settings, \emph{correlational information} can be of crucial value. Specifically, suppose the robot is equipped with a prior about the relative spatial locations of object types (\eg, stoves tend to be near pepper shakers). Then, it can leverage this information as a powerful heuristic to narrow down or ``focus'' the search space, by first locating easier-to-detect objects that are highly correlated with the target object (Fig.~\ref{fig:teaser}). Doing so has the potential to greatly improve search efficiency; unfortunately, previous approaches to object search with correlational information tend to resort to ad-hoc or greedy search strategies~\cite{aydemir2013active,kollar2009utilizing,zeng2020semantic,wixson1994using} or assemble a collection of independent components \cite{aydemir2011search}, which may not scale well to complex environments.  

We follow a long line of work that models the object search problem as a partially observable Markov decision process (POMDP) \cite{aydemir2013active,li2016act,xiao_icra_2019,zheng2020multi}. This formalization is useful because object search over long horizons is naturally a sequential, partially observed decision-making problem: the robot must (1) search for the target object by visiting multiple viewpoints in the environment sequentially, and (2) maintain and update a measure of uncertainty over the location of the target object, via its belief state. However, existing POMDP-based approaches assume object independence for scalability of maintaining and reasoning about the belief states and do not consider correlational information between objects in the environment during the search process~\cite{zheng2020multi,wandzel2019multi,holzherr2021efficient}.

We introduce COS-POMDP (Correlational Object Search POMDP), a general planning framework for optimal object search with given correlational information. Critically, COS-POMDPs avoid the exponential blow-up of naively maintaining a joint belief about all objects while preserving optimal solutions to this exponential formulation. COS-POMDPs model correlational information using a correlation-based observation model. The correlational information is given to the robot as a factored joint distribution over object locations. In practice, this distribution can be approximated
by learning it from data \cite{kollar2009utilizing,krishna2017visual} or interpreting human speech~\cite{kollar2010toward, sloop-roman-2020}.
We address scalability by proposing a hierarchical
planning algorithm, where a high-level COS-POMDP plans subgoals, each fulfilled by a low-level planner that plans with low-level actions (\ie, given primitive actions); both levels plan online based on a shared and updated COS-POMDP belief state, enabling efficient closed-loop planning.

We evaluate the proposed approach in AI2-THOR \cite{kolve2017ai2}, a realistic simulator of household environments, and we use YOLOv5 \cite{redmon2016you,glenn_jocher_2020_4154370} as the object detector. Our results show that, when the given correlational information is accurate, COS-POMDP leads to more robust search perfomance for target objects that are hard-to-detect. In particular, for target objects with a true positive detection rate below 40\%, COS-POMDP significantly outperforms the POMDP baseline not using correlational information by 42.1\% and a greedy, next-best view baseline \cite{zeng2020semantic} by 210\% in terms of SPL (success weighted by inverse path length) \cite{anderson2018evaluation}, a recently developed metric that reflects both search success and efficiency. 

\section{Related Work}
Object search involves a wide range of subproblems (\eg, perception ~\cite{aydemir2013active,xiao_icra_2019}, planning ~\cite{ye1996sensor,li2016act}, manipulation \cite{wong2013manipulation,novkovic2020object}) and different types of target objects (moving \cite{brown1980optimal} or static \cite{ye1996sensor}).
We consider static objects and an environment where the set of possible object locations is given, but we assume no object location is known a priori.

\citet{garvey1976} and \citet{wixson1994using} pioneered the paradigm of \emph{indirect search}, where an intermediate object (such as a desk) that is typically easier to detect is located first, before the target object (such as a keyboard). More recently, probabilistic graphical models have been used to model object-room or object-object spatial correlations \cite{zeng2020semantic,aydemir2013active,kollar2009utilizing,lorbach2014prior,amiri2022reasoning}. In particular, \citet{zeng2020semantic}~proposed a factor graph representation for different types of object spatial relations. Their approach produces search strategies in a greedy fashion by selecting the next-best view to navigate towards, based on a hybrid utility of navigation cost and the likelihood of detecting objects. In our evaluation, we compare our sequential decision-making approach with a greedy, next-best view baseline based on that work \cite{zeng2020semantic}.

Recently, the problem of semantic visual navigation \cite{zhu2017target,batra2020objectnav,Wortsman_2019_CVPR,qiu2020learning,mayo2021visual} received a surge of interest in the deep learning community.
In this problem, an embodied agent is placed in an unknown environment and tasked to navigate towards a given semantic target (such as ``kitchen'' or ``chair''). The agent typically has access to behavioral datasets for training on the order of millions of frames and the challenge is typically in generalization. Our work considers the standard evaluation metric (SPL \cite{anderson2018evaluation}) and task success criteria (object visibility and distance threshold \cite{batra2020objectnav}) from this body of work. However, our setting differs fundamentally in that the search strategy is not a result of training but a result of solving an optimization problem.

Finally, our hierarchical planning algorithm for COS-POMDPs differs in not limiting POMDP to local use \cite{aydemir2011search}, or assuming navigation tasks for low-level macro-actions \cite{lee21magic}.

\section{Problem Formulation}~\label{sec:prob}
We present a general formulation of correlational object search as a planning problem, where a robot must search for a target object given correlational information with other objects in the environment. We begin by describing the underlying search environment and the capabilities of the robot, followed by the problem definition.
\subsection{Search Environment and Robot Capabilities}
\label{sec:cos_envrob}
The search environment contains a target object and $n$ additional static objects. The set of possible object locations is discrete, denoted as $\mc{X}$. The locations of the target object $\xtarget\in\mc{X}$ and other objects $x_1,\ldots, x_n\in \mc{X}$ are unknown to the robot and follow a latent joint distribution $\Pr(x_1,\ldots,x_n,\xtarget)$. The robot is given as input a factored form of this distribution, defined later in Sec.~\ref{sec:cos}.

The robot can observe the environment from a discrete set of viewpoints, where each viewpoint is specified by the position and orientation of the robot's camera. These viewpoints form the necessary state space of the robot, denoted as $\Srobot$. The initial viewpoint is denoted as $\srobotinit$. By taking a primitive move action $a$ from the set $\mc{A}_m$, the robot changes its viewpoint subject to transition uncertainty $T_m(\srobot',\srobot,a)=\Pr(\srobot'|\srobot,a)$. Also, the robot can decide to finish a task at any timestep by choosing a special action \DECLARE, which deterministically terminates the process.

At each timestep, the robot receives an observation $z$ factored into two independent components $z=(\zrobot,\zobjects)$. The first component $\zrobot\in\Srobot$ is an estimation of the robot's current viewpoint following the observation model $\Orobot(\zrobot,\srobot)=\Pr(\zrobot|\srobot)$. The second component $\zobjects=(z_1,\ldots, z_n,\ztarget)$ is the result of performing object detection. Each element, $z_i\in\mc{X}\cup\{\texttt{null}\}$, $i\in\{1,\ldots,n,\text{target}\}$, is the detected location of object $i$ within the field of view, or \texttt{null} if not detected. The observation $z_i$ about object $i$ is subject to limited field of view and sensing uncertainty captured by a \emph{detection model} $D_i(z_i,x_i,\srobot)=\Pr(z_i|x_i,\srobot)$; Here, a common conditional independence assumption in object search is made \cite{zeng2020semantic,wandzel2019multi}, where $z_i$ is conditionally independent of the observations and locations of all other objects given its location and the robot state $\srobot$. The set of detection models for all objects is $\mc{D}=\{D_1,\ldots,D_n,D_{\text{target}}\}$. In our experiments, we obtain parameters for the detection models based on the performance of the vision-based object detector (Sec.~\ref{sec:exp:detectors}). 

\subsection{The Correlational Object Search Problem}\label{sec:cos}
Although the joint distribution of object locations is latent, the robot is assumed to have access to a factored form of that distribution, that is, $n$ conditional distributions, $\mc{C}=\{C_1,\ldots,C_n\}$ where $C_i(x_i,\xtarget)=\Pr(x_i|\xtarget)$ specifies the spatial correlation between the target and object $i$. We call each $C_i$ a \emph{correlation model}. This model can be learned from data or, in our case, be given by a domain expert.

Formally, we define the \emph{correlational object search} problem as follows. Given as input a tuple $(\mc{X},\mc{C},\mc{D},\srobotinit, \Srobot,$ $\Orobot,\mc{A}_m,T_m)$, the robot must perform a sequence of actions, $a_1,\ldots,a_t$, where $a_1,\ldots,a_{t-1}\in\mc{A}_m$ and the last action is \DECLARE. The success criteria depends on the robot state and the target location at the time of \DECLARE, and the robot should minimize the distance traveled to find the object. In our evaluation in AI2-THOR, we use the success criteria recommended by \citet{batra2020objectnav}, defined in Sec.~\ref{sec:ai2thor}.

\section{Correlational Object Search as a POMDP}
In this
section, we introduce the COS-POMDP, a POMDP formulation that addresses the
correlational object search problem, followed by a discussion on its optimality.
We begin with a brief review of POMDPs \cite{kaelbling1998planning,shani2013survey,kurniawati2022partially}.

\subsection{Background: POMDPs}\label{sec:pomdps}
A POMDP is formally defined as a tuple $(\mc{S},\mc{A},\mc{Z},$  $T,O,R,\gamma)$, where $\mc{S},\mc{A},\mc{Z}$ denote the state, action, and observation spaces, $T(s',a,s)=\Pr(s'|s,a)$, $O(z, s',a)=\Pr(z|s',a)$ are the transition and observation models, and $R(s,a)\in\mathbb{R}$ is the reward function. At each timestep, the agent takes
an action $a\in\mc{A}$, the environment state transitions from $s\in\mc{S}$ to $s'\in\mc{S}$ according to $T$, and the
agent receives an observation $z\in\mc{Z}$ from the environment according to $O$.

The agent typically maintains a \emph{belief state} $b_t:\mc{S}\rightarrow[0,1]$, a
distribution over the states and a sufficient statistic for the history of past
actions and observations $h_t=(az)_{1:t-1}$. The agent updates its
belief 
after taking action $a$ and receiving observation $z$:
$b^{z,a}(s')=\eta O(z, s',a)\sum_{s}T(s',a,s)b(s)$, where $\eta$ is the
normalizing constant
\cite{kaelbling1998planning}. 

The solution to a POMDP is a \emph{policy}
$\pi$ that maps a history to an action.  The \emph{value} of a POMDP at a history
under policy
$\pi$ is the expected discounted cumulative reward following that policy:
$V_\pi(h)=\E[\sum_{t=0}^\infty \gamma^t R(s_t,\pi(h_t)|h_0=h]$ where $\gamma$ is the discount factor. The optimal value at the history is
$V(h)=\max_{\pi}V_\pi(h)$.

\subsection{COS-POMDP Definition}
\label{sec:cospomdps}

Given an instance of the correlational object search problem defined in Sec.~\ref{sec:cos}, we define the Correlational Object Search POMDP (COS-POMDP) as follows:
\begin{itemize}
\item\textbf{State space.} The state space $\mc{S}$ is factored to include the robot state $\srobot\in\Srobot$ and the target state $\xtarget\in\mc{X}$. A state $s\in\mc{S}$ can be written as $s=(\srobot,\xtarget)$. Importantly, no other object state is included in $\mc{S}$.

\item\textbf{Action space.}
 The action space is $\mc{A}=\mc{A}_m\cup\{\DECLARE\}$.

\item\textbf{Observation space.}
The observation space $\mc{Z}$ is factored over the objects, and each $z\in \mc{Z}$ is written as $z=(\zrobot,\zobjects)$, where $\zobjects=(z_1,\ldots,z_n,\ztarget)$.

\item\textbf{Transition model.} The objects are assumed to be static.  Actions
$a_m\in\mc{A}_m$ change the robot state from $\srobot$ to $\srobot'$ according to $T_m$, and taking the
$\DECLARE$ action terminates the task deterministically.

\item\textbf{Observation model.} By definition of $z$, $\Pr(z|s)=\Pr(\zrobot|\srobot)\Pr(\zobjects|s)$ where $\Pr(\zrobot|\srobot)$ is defined by $\Orobot(\zrobot,\srobot)$. 
Under the conditional independence
assumption in Sec.~\ref{sec:prob}, $\Pr(\zobjects|s)$ can be compactly factored:
\begin{align}
&\Pr(\zobjects|s)=\Pr(z_1,\ldots,z_n,\ztarget|\xtarget,\srobot)\\
&\ \ =\Pr(\ztarget|\xtarget,\srobot)\prod_{i=1}^n\Pr(z_i|\xtarget,\srobot)\label{eq:factoring}
\end{align}

The first term in Eq~(\ref{eq:factoring}) is defined by $\Dtarget$, and each $\Pr(z_i|\xtarget,\srobot)$ is called a \emph{correlational observation model},
written as:
\begin{align}
  &\Pr(z_i|\xtarget,\srobot)=\sum_{x_i\in\mc{X}}\Pr(x_i,z_i|\xtarget,\srobot)\\
  &\qquad=\sum_{x_i\in\mc{X}}\Pr(z_i|x_i,\srobot)\Pr(x_i|\xtarget)\label{eq:corr_obz_model}
\end{align}
where the two terms in Eq~(\ref{eq:corr_obz_model}) are the detection model $D_i\in\mc{D}$ and correlation model $C_i\in\mc{C}$, respectively.

\item\textbf{Reward function.} The reward function, $R(s,a)=R(\srobot,\xtarget,a)$, is defined as follows. Upon taking \DECLARE, the task outcome is determined based on $\srobot, \xtarget$,
which is successful if the robot orientation is facing the target and its position is within a distance threshold to the target.
If successful, then the robot receives $\rmax\gg 0$, and $\rmin\ll 0$ otherwise.
Taking a move action from $\mc{A}_m$ receives a negative reward which corresponds to the action's cost. In our experiments, we set $\rmax=100$ and $\rmin=-100$. Each primitive move action (\eg, \texttt{MoveAhead})  receives a step cost of $-1$.
\end{itemize}

\subsection{Optimality of COS-POMDPs}
The state space of a COS-POMDP involves only the robot and target object states. A natural question arises: have we lost any necessary information? In this section, we show that
COS-POMDPs are optimal, in the following sense: if we imagine solving a ``full'' POMDP corresponding  to the COS-POMDP, whose state space contains all object states, then the solutions to the COS-POMDP are equivalent. Note that a belief state in this ``full'' POMDP scales exponentially in the number of objects.

We begin by precisely defining the ``full'' POMDP, henceforth called
the F-POMDP, corresponding to a COS-POMDP. The F-POMDP has identical action space, observation space, and transition model as the COS-POMDP. The reward function is also identical since it only depends on the target object state, robot state, and the action taken. F-POMDP differs in the state space and observation model:
\begin{itemize}
\item \textbf{State space.} The state is $s=(\srobot,\xtarget,x_1,\ldots,x_n)$.
\item \textbf{Observation model.} Under the conditional independence assumption stated in Sec.~\ref{sec:prob}, the model for observation $z_i$ of object $x_i$ involves just the detection model: $\Pr(z_i| s)=\Pr(z_i| x_i, \srobot)$.
  \end{itemize}

  Since the COS-POMDP and the F-POMDP share the same action and observation spaces, they have the same history space as well. We first show that given the same policy, the two models have the same distribution over histories.\vspace{0.05in}

  \textbf{Theorem 1.}
  Given any policy $\pi:h_t\rightarrow a$, the distribution of histories is identical between the COS-POMDP and the F-POMDP.

  \begin{proof} (Sketch)
    We prove this by induction. When $t=1$, the statement is true because both histories are empty. The inductive hypothesis assumes that the distributions $\Pr(h_t)$ is the same for the two POMDPs at $t\geq 1$. Then, by definition, $\Pr(h_{t+1})=\Pr(h_t,a_t,z_t)=\Pr(z_t|h_t,a_t)\Pr(a_t|h_t)\Pr(h_t)$. Note that $\Pr(a_t|h_t)$ is the same under the given $\pi$. We can show the two POMDPs have the same $\Pr(z_t|h_t,a_t)$; the full proof is available in Appendix \ref{appdx:proof1}.\footnote{The appendix is available at \url{https://arxiv.org/pdf/2110.09991.pdf}}
  \end{proof}\vspace{0.05in} 

    Using Theorem 1, we are equipped to make a statement
    about the value of following a given policy in either the COS-POMDP or the
    F-POMDP.\vspace{0.05in}

    \textbf{Corollary 1.} Given any policy $\pi:h_t\rightarrow a$ and history $h_t$, the value $V_{\pi}(h_t)$ is identical between the COS-POMDP and the F-POMDP.

    \begin{proof}
      By definition, the value of a POMDP at a history is the expected discounted cumulative reward with respect to the distribution of future action-observation pairs. Theorem~1 states that the COS-POMDP and F-POMDP have the same distribution of histories given $\pi$. Furthermore, the reward function depends only on the states of the robot and the target object. Thus, this expectation is equal for the two POMDPs at any $h$.
    \end{proof}\vspace{0.05in}

  Finally, we can show that COS-POMDPs are optimal in the sense that we described before.\vspace{0.05in}

 \textbf{Corollary 2.}
An optimal policy $\pi^*$ for either the COS-POMDP or the F-POMDP is also optimal for the other.

\begin{proof}
  Suppose, without loss of generality, that $\pi^*$ is optimal for the COS-POMDP but not the F-POMDP. Let $\pi'$ be the optimal policy for the F-POMDP. By the definition of optimality, for at least some history $h$ we must have $V_{\pi'}(h) > V_{\pi^*}(h)$. By Corollary 1, for any such $h$ the COS-POMDP also has value $V_{\pi'}(h)$, meaning $\pi^*$ is not actually optimal for the COS-POMDP; this is a contradiction.
\end{proof}

\section{Hierarchical Planning}
\label{sec:hier}
Despite the optimality-preserving reduction of state space in a COS-POMDP, directly planning over the primitive move actions is not scalable to practical domains even for state-of-the-art online POMDP solvers \cite{kurniawati2022partially}.
Nevertheless, planning POMDP actions at the primitive level has the benefit of controlling fine-grained movements, allowing goal-directed behavior to emerge automatically at this level. Therefore, we seek an algorithm that can reason about both searching over a large region and searching carefully in a local region. 

To this end, we propose a hierarchical planning algorithm to apply COS-POMDPs to realistic domains (Fig.~\ref{fig:method}). The pseudocode and a detailed description is provided in Appendix \ref{appendix:hier}. As an overview: (1) A topological graph is first dynamically generated to reflect the robot's belief in the target location. Nodes are places accessible by the robot, and edges indicate navigability between places \cite{toponets-iros-2019}. (2)~Then, a high-level COS-POMDP is instantiated which plans \emph{subgoals} that can be either navigating to another place, or searching locally at the current place. Both types of subgoals can be understood as viewpoint-changing actions, except the latter keeps the viewpoint the same. (3) At each timestep, a subgoal is planned using a POMDP solver, and a low-level planner is instantiated corresponding to the subgoal. This low-level planner then plans to output an action from the action set  $\mc{A}=\mc{A}_m \cup \{\DECLARE\}$, which is used for execution. In our implementation, for navigation subgoals, an A$^*$ planner is used, and for the subgoal of searching locally, a low-level COS-POMDP is instantiated whose action space is the primitive movements $\mc{A}_m$, and solved using a POMDP planner~\cite{silver2010monte}. (4) Upon executing the low-level action, the robot receives an observation from its on-board object detector. This observation is used to update the belief of both the high-level COS-POMDP as well as the low-level COS-POMDP (if it exists). (5) If the cumulative belief captured by the nodes in the current topological graph is below a threshold (50\% in our experiments), then the topological graph is regenerated to better reflect the belief. (6) Finally, the process starts over from step (3). If the high-level COS-POMDP plans a new subgoal different from the current one, then the low-level planner is re-instantiated. Our algorithm plans actions for execution in an online, closed-loop manner, allowing for reasoning about viewpoint changes at the level of both places in a topological graph and fine-grained movements. This is efficient in practice because typical mobile robots can be controlled both at the low level of motor velocities and the high level of navigation goals \cite{macenski2020marathon,zheng-ros-navguide}. 



\begin{figure}[t]
    \centering
    \includegraphics[width=\linewidth]{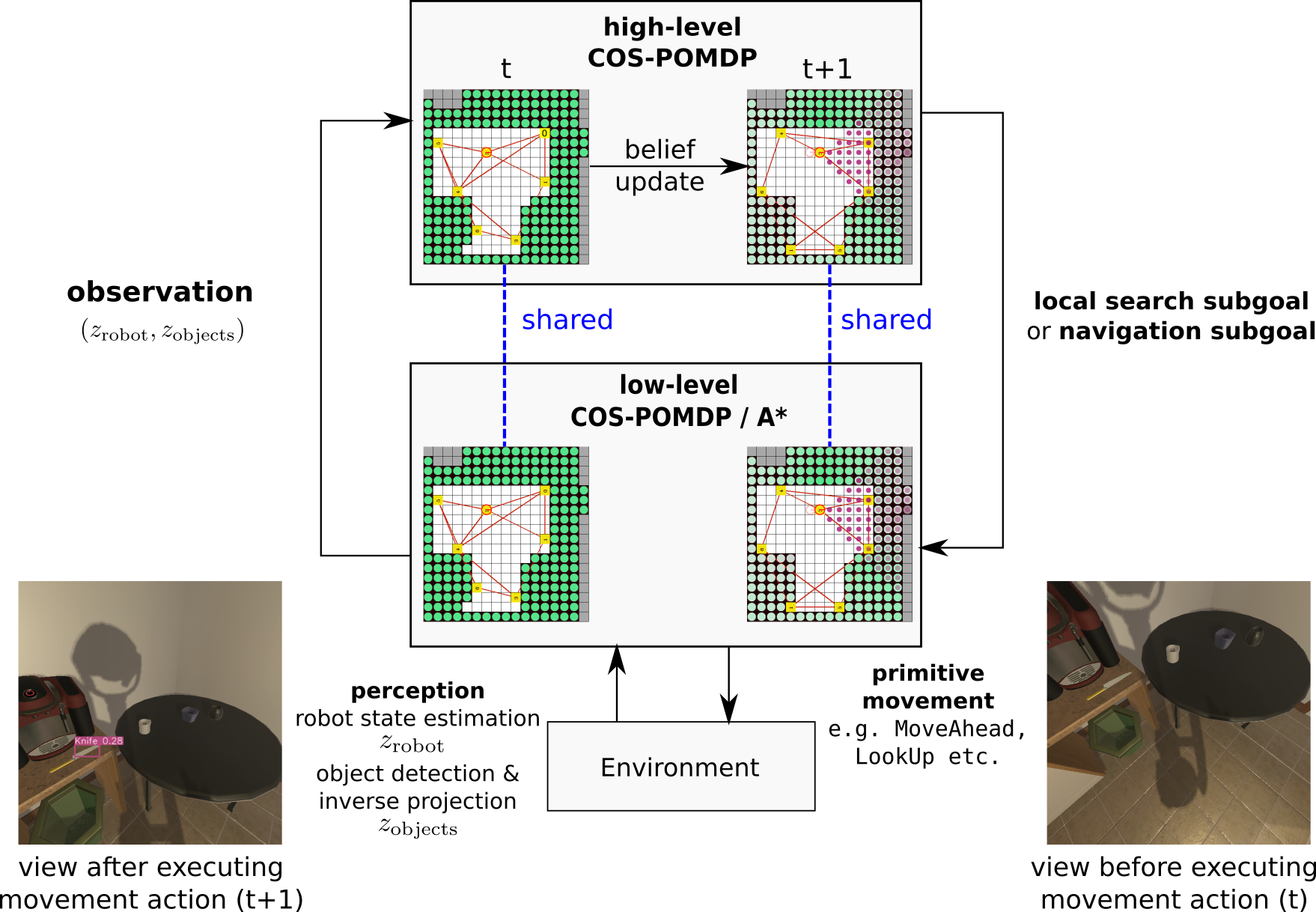}
    \caption{\textbf{Illustration of the Hierarchical Planning Algorithm.} A high-level COS-POMDP plans subgoals fed into a low-level planner that produces low-level actions. The belief state is shared across the levels. Both levels plan with updated beliefs at every timestep.
    }
    \label{fig:method}
    \vspace{-0.15in}
\end{figure}

\section{Experimental Setup}

We test the following hypotheses through our experiments: (1) Leveraging correlational information with easier-to-detect objects can benefit the search for hard-to-detect objects; (2)~Optimizing over an action sequence improves performance compared to greedily choosing the next-best view.

\subsection{AI2-THOR}
\label{sec:ai2thor}
We conduct experiments in AI2-THOR \cite{kolve2017ai2}, a realistic simulator of in-household rooms. It has a total of 120 scenes divided evenly into four room types: \emph{Bathroom}, \emph{Bedroom},
\emph{Kitchen}, and \emph{Living room}.
For each room type, we use the first 20 scenes for training a vision-based object detector and learning object correlation models (used in some experiments), and the last 10 scenes for validation. 

The robot can take primitive move actions from the set:  $\{\texttt{MoveAhead},
\texttt{RotateLeft},$ $\texttt{RotateRight},$ $\texttt{LookUp},$ $\texttt{LookDown}\}$. \texttt{MoveAhead} moves
the robot forward by 0.25m. \texttt{RotateLeft}, \texttt{RotateRight} rotate the robot
in place by 45$^\circ$. \texttt{LookUp}, \texttt{LookDown} tilt the camera up or down by 30$^\circ$. The robot observes the pose of its current viewpoint without noise. To be successful, when the robot takes \texttt{Done}, the robot must be within a Euclidean distance of 1.0m from the target object while the target object is visible in the camera frame. The maximum number of steps allowed is 100.

\subsection{Object Detector}\label{sec:exp:detectors}
We use YOLOv5 \cite{glenn_jocher_2020_4154370}, a popular vision-based object detector. This contrasts previous works evaluated using a ground truth object
detector \cite{qiu2020learning} or detectors with synthetic noise and detection
ranges \cite{zeng2020semantic,holzherr2021efficient}.
We collect training data by randomly placing the robot in
the training scenes.




\begin{figure}[t]
    \centering
    \includegraphics[width=\linewidth]{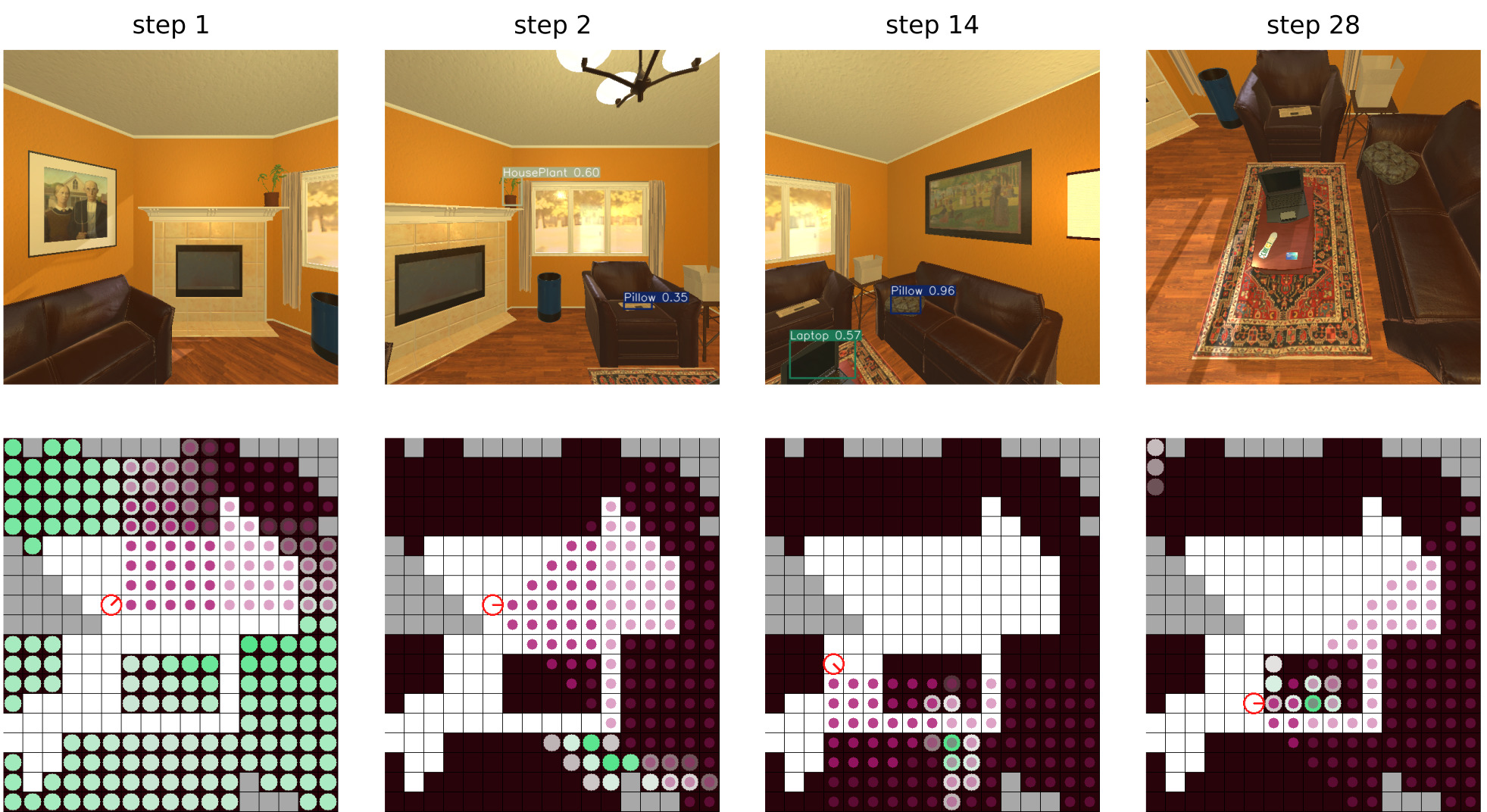}
    \caption{\textbf{Example Sequence.} Top: first-person view with object detection bounding boxes. Bottom: Visualization of belief state corresponding to each view. See Fig.~\ref{fig:method} for the legend of the belief state visualization. Our method (COS-POMDP) successfully finds a \texttt{CreditCard} in a living room scene, leveraging the detection of other objects such as \texttt{FloorLamp} and \texttt{Laptop}. For more examples, please refer to the video at \url{https://youtu.be/RneTq4o0a-A}.}
    \label{fig:example}
    \vspace{-0.05in}
\end{figure}
\label{sec:experiments}
\begin{figure}[t]
    \centering
    \includegraphics[width=0.95\linewidth]{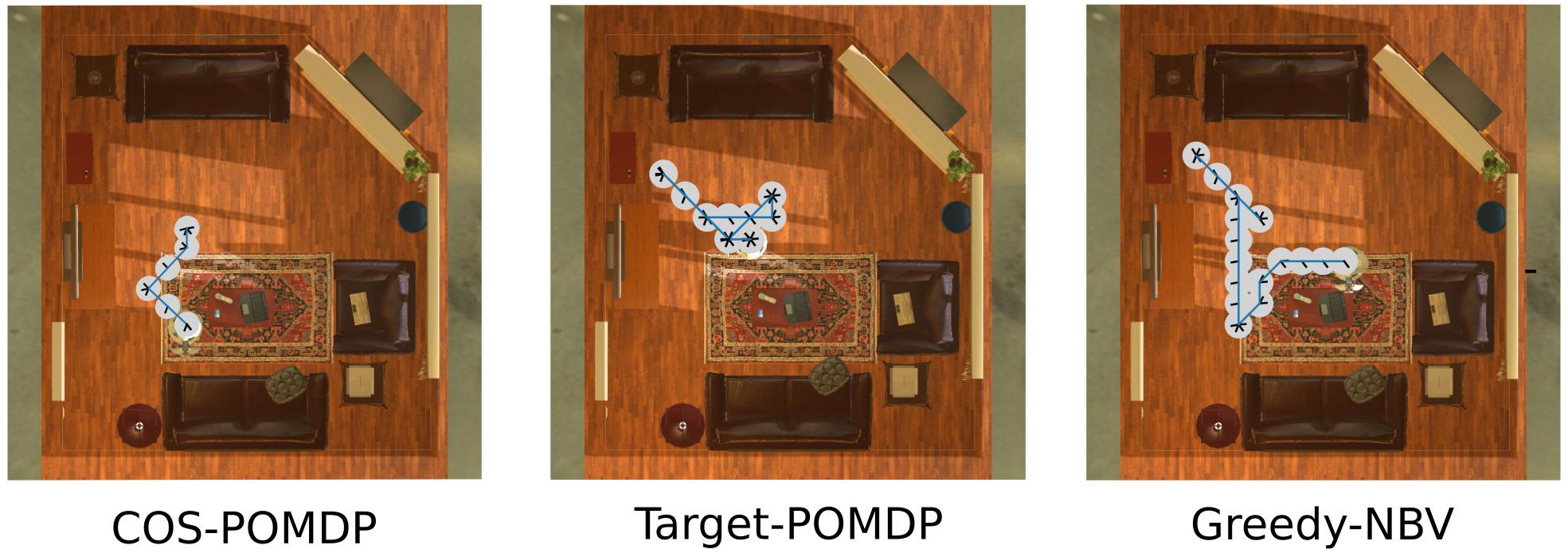}
    \caption{
    Visualization of robot trajectory produced by different methods for the example shown in Fig.~\ref{fig:example}. A gray circle combined with a black line segment indicates a viewpoint.
    }
    \label{fig:paths}
    \vspace{-0.2in}
  \end{figure}

\textbf{Detection Model.} Since vision detectors can sometimes detect small objects from far away, we consider a line-of-sight detection model with a limited field of view angle:
\begin{align*}
  &D(z_i,x_i,\srobot)=\Pr(z_i|x_i,\srobot)\\
  &\qquad=\begin{cases}
    1.0 - \text{TP} & s_i\in \mc{V}(\srobot) \land z_i = \texttt{null}\\
    \delta \text{FP} / |\mc{V}_{E}(r)| & s_i\in \mc{V}(\srobot) \land \norm{z_i- x_i} > 3\sigma\\
    \delta \mc{N}(z_i;x_i,\sigma^2) & s_i\in \mc{V}(\srobot) \land \norm{z_i- x_i}\leq 3\sigma\\
    1.0 - \text{FP} & s_i\not\in \mc{V}(\srobot) \land z_i = \texttt{null}\\
    \delta \text{FP} / |\mc{V}_{E}(r)|  & s_i\not\in \mc{V}(\srobot) \land z_i \neq \texttt{null}\\
  \end{cases}
\end{align*}
This detection model is parameterized by: TP, the true positive rate; FP, the false positive
rate; $r$, the average distance between the robot and the object for true positive detections;
$\sigma$, the width of a small region around the true object location where a detection
made within that region, though not exactly accurate, is still accepted as a true positive detection. We set $\sigma=0.5$m. The notation $\mc{N}(\cdot)$ denotes a Gaussian distribution. The $\mc{V}(\srobot)$ denotes the line-of-sight field of view with a 90$^\circ$ angle. The $\mc{V}_E(r)$ denotes the region inside the field of view that is within distance $r$ from the robot.
The weight $\delta=1$ if the detection is within $\mc{V}_E(r)$, and otherwise $\delta=\exp(-\norm{z_i-\srobot}-r)^2$.

\subsection{Target Objects}
We choose the target and correlated object classes based on detection
statistics. The list of target object classes and other correlated classes for
each room type is listed below (in no particular order). For detection
statistics, please refer to Table~\ref{tab:main} and Table~\ref{tab:detcorr}
(Appendix~\ref{appdx:detcorr}).
\begin{itemize}
\item \textbf{Bathroom.} \emph{Targets} are \texttt{Fauct}, \texttt{Candle}, \texttt{ScrubBrush};
  \emph{Correlated} objects are  \texttt{ToiletPaperHanger}, \texttt{Towel}, \texttt{Mirror}, \texttt{Toilet}, \texttt{SoapBar}.
\item \textbf{Bedroom.} \emph{Targets} are  \texttt{AlarmClock}, \texttt{CellPhone}, \texttt{Book};
  \emph{Correlated objects} are  \texttt{Laptop}, \texttt{Bed}, \texttt{DeskLamp}, \texttt{Mirror}, \texttt{LightSwitch}.
\item \textbf{Kitchen.} \emph{Targets}: \texttt{Bowl}, \texttt{Knife}, \texttt{PepperShaker};
\emph{Correlated objects} are \texttt{Lettuce}, \texttt{LightSwitch}, \texttt{Microwave}, \texttt{Plate}, \texttt{StoveKnob}
\item \textbf{Living room.} \emph{Targets} are \texttt{CreditCard}, \texttt{RemoteControl}, \texttt{Television};
\emph{Correlated objects} are   \texttt{Pillow}, \texttt{Laptop}, \texttt{LightSwitch}, \texttt{HousePlant},  \texttt{FloorLamp}, \texttt{Painting}.
\end{itemize}

\subsection{Correlation Model}
We consider a binary correlation model that takes into account whether the correlated object and the target are close or far. Note that our method is not specific to this model. We use this model since it is applicable between arbitrary object classes and can be easily estimated based on object instances.
\begin{align}
  &C(\xtarget,x_i)=\Pr(x_i|\xtarget)\\
  &=\begin{cases}
    1 & \text{Close}(i,\text{target}) \land \norm{x_i - \xtarget} < d(i,\target)\\
    0 & \text{Close}(i,\text{target}) \land \norm{x_i - \xtarget} \geq d(i,\target)\\
    1 & \text{Far}(i,\text{target}) \land \norm{x_i - \xtarget} > d(i,\target)\\
    0 & \text{Far}(i,\text{target}) \land \norm{x_i - \xtarget} \leq d(i,\target)\\
    \end{cases}
\end{align}
where Close$(\cdot,\cdot)$ and Far$(\cdot,\cdot)$ are opposite, class-level predicates, $\norm{\cdot}$ denotes the Euclidean distance, and $d(\cdot,\cdot)$ is the expected distance between the two object classes. In Sec.~\ref{sec:results}, we conduct an ablation study where $d(\cdot,$target$)$ is estimated under different scenarios: \textbf{accurate}: based on object ground truth locations in the deployed scene; \textbf{estimated} (est): based on instances in training scenes; \textbf{wrong} (wrg): same as accurate except we flip the close/far relationship between the objects so that they do not match the scene.

\begin{table*}[!t]
    \setlength{\tabcolsep}{2pt}
    \renewcommand{\arraystretch}{1.0}{
  \resizebox{\textwidth}{!}{%
    \centering

\begin{tabular}{l|ccc|ccc|ccc|ccc}
\specialrule{.8pt}{2pt}{0pt}
                               & \multicolumn{3}{c|}{Bathroom}                                                               & \multicolumn{3}{c|}{Bedroom}                          & \multicolumn{3}{c|}{Kitchen}                         & \multicolumn{3}{c}{Living room}                                                                                                                         \\
Method                    & SPL (\%)           & DR                 & SR (\%)    & SPL (\%)           & DR                  & SR (\%)    & SPL (\%)           & DR                 & SR (\%)    & SPL (\%)           & DR                  & SR (\%)    \\
\specialrule{.4pt}{0pt}{0pt}
Random                         & 0.00 (0.00)        & -82.75 (3.43)      & 0.00       & 0.00 (0.00)        & -81.51 (3.33)       & 0.00       & 6.90 (9.81)        & -68.51 (15.61)     & 6.90       & 0.00 (0.00)        & -82.37 (3.62)       & 0.00       \\
Greedy-NBV                     & 14.34 (9.12)       & -19.86 (11.87)     & 34.48      & 16.92 (11.70)      & -17.52 (7.32)       & 26.67      & 11.61 (8.72)       & -17.60 (12.41)     & 31.03      & 7.13 (7.11)        & -21.41 (8.21)       & 20.00      \\
Target-POMDP                   & 19.88 (9.47)       & \bb{-7.37 (12.42)} & \bb{55.17} & 19.79 (12.81)      & -20.79 (11.29)      & 26.67      & 13.80 (8.67)       & -20.17 (12.83)     & 34.48      & 24.36 (13.28)      & -33.58 (11.88)      & 40.00      \\
COS-POMDP                      & \bb{30.64 (12.73)} & -14.48 (11.58)     & \bb{55.17} & \bb{24.76 (12.95)} & \bb{-15.57 (9.16)}  & \bb{40.00} & \bb{20.45 (12.00)} & \bb{-6.55 (12.73)} & \bb{41.38} & \bb{24.99 (13.95)} & \bb{-14.08 (14.22)} & \bb{43.33} \\
\specialrule{.4pt}{0.7pt}{0.9pt}
COS-POMDP (gt)                 & \bb{31.08 (13.31)} & -13.47 (12.67)     & \bb{58.62} & \bb{26.67 (13.13)} & \bb{-11.09 (12.07}) & 40.00      & \bb{35.58 (13.30)} & \bb{-2.75 (14.37)} & \bb{62.07} & \bb{32.88 (14.25)} & \bb{-13.81 (13.22)} & \bb{56.67} \\
\specialrule{.4pt}{0.7pt}{0.7pt}
COS-POMDP (est)                & 17.20 (10.21)      & -20.96 (10.75)     & 41.38      & 16.78 (11.68)      & -31.60 (10.05)      & 30.00      & 8.39 (7.94)        & -31.36 (13.42)     & 20.69      & 14.07 (10.65)      & -43.76 (13.30)      & 26.67      \\
COS-POMDP (wrg)                & 11.89 (8.14)       & -16.55 (10.23)     & 27.59      & 14.70 (10.92)      & -17.33 (8.38)       & 23.33      & 10.51 (8.02)       & -20.68 (10.40)     & 27.59      & \textbf{31.41 (14.50) }     & -15.94 (9.45)       & \textbf{53.33}      \\
\bottomrule
\end{tabular}
}
  }
  \caption{\textbf{Main and Ablation Study Results.} Unless otherwise specified, all methods use the YOLOv5 \cite{glenn_jocher_2020_4154370} vision detector and are given accurate correlational information. \emph{Target-POMDP} uses hierarchical planning but only the target object is detectable. \emph{Greedy-NBV} is a next-best view approach based on \cite{zeng2020semantic}. \emph{Random} chooses actions uniformly at random. The highest value of each metric per room type is bolded. Parentheses contain 95\% confidence interval. Ablation study results are bolded if it outperforms the best result from the main evaluation.
  \emph{COS-POMDP} is more consistent, performing either the best or the second best across all room types and metrics.
  }
  \label{tab:main}
  \vspace{-0.05in}
\end{table*}

\begin{table*}[!t]
\setlength{\tabcolsep}{4pt}
\renewcommand{\arraystretch}{1.2}{
  \resizebox{\textwidth}{!}{%
    \centering

\begin{tabular}{ll|ccc|ccc|ccc|ccc}
\specialrule{.8pt}{2pt}{0pt}
                              &               &      &      &         & \multicolumn{3}{c|}{Greedy-NBV } & \multicolumn{3}{c|}{Target-POMDP}       & \multicolumn{3}{c}{COS-POMDP}                     \\
        Room Type             & Target Class  & TP   & FP   & $r$ (m) & SPL (\%)      & DR               & SR (\%) & SPL (\%)      & DR             & SR (\%) & SPL (\%)      & DR             & SR (\%) \\
\specialrule{.4pt}{0pt}{0pt}
  \multirow{3}{*}{Bathroom}   & Faucet        & 56.1 & 8.0  & 2.16    & 28.31 (19.58) & 0.73 (22.10)     & \bb{70.00}   & 34.67 (22.86) & \bb{8.00 (24.67)}   & \bb{70.00}   & 28.18 (27.25) & -23.27 (24.36) & 50.00   \\
                              & Candle        & 29.4 & 2.4  & 1.81    & 12.52 (20.12) & -22.81 (20.80)   & 22.22   & 16.56 (13.36) & -7.98 (28.99)  & \bb{66.67}   & \bb{33.89 (21.83)} & \bb{-2.94 (19.08)}  & \bb{66.67}   \\
                              & ScrubBrush    & 64.3 & 9.9  & 1.71    & 2.00 (4.52)   & -37.79 (17.36)   & 10.00   & 8.09 (10.79)  & -22.18 (13.51) & 30.00   & \bb{30.18 (25.78)} & \bb{-16.07 (22.13)} & \bb{50.00}   \\

\cline{1-14}
\multirow{3}{*}{Bedroom}      & AlarmClock    & 79.6 & 7.4  & 2.77    & \bb{39.49 (31.18)} & \bb{-5.54 (18.07)}    & \bb{50.00}   & 14.31 (22.01) & -23.78 (14.43) & 20.00   & 31.57 (30.85) & -15.85 (21.03) & 40.00   \\
                              & Book          & 62.6 & 4.9  & 2.05    & 8.42 (12.72)  & -20.10 (11.71) & 20.00   & \bb{29.70 (28.85)} & -13.94 (27.69) & 40.00   & 25.92 (22.50) & \bb{-12.56 (16.69)} & \bb{50.00}   \\
                              & CellPhone     & 50.0 & 3.9  & 1.69    & 2.85 (6.44)   & -26.91 (5.88)    & 10.00   & 15.36 (23.21) & -24.64 (22.20) & 20.00   & \bb{16.80 (21.48)} & \bb{-18.29 (16.16)} & \bb{30.00}   \\
\cline{1-14}
\multirow{3}{*}{Kitchen}      & Bowl          & 60.6 & 11.5 & 1.75    & 19.88 (26.57) & -15.76 (32.76)   & 33.33   & 16.33 (16.00) & -10.06 (27.39) & \bb{55.56}   & \bb{20.37 (20.70)} & \bb{-3.33 (27.27)}  & 44.44   \\
                              & Knife         & 37.7 & 8.7  & 1.68    & 7.40 (11.42)  & -18.94 (23.71)   & 30.00   & 4.62 (10.45)  & -36.36 (15.51) & 10.00   & \bb{23.97 (25.58)} & \bb{-2.59 (25.33)}  & \bb{50.00}   \\
                              & PepperShaker  & 38.1 & 9.4  & 1.43    & 8.39 (10.53)  & -17.90 (17.39)   & 30.00   & \bb{20.69 (21.03)} & \bb{-13.07 (27.64)} & \bb{40.00}   & 17.01 (24.19) & -13.41 (22.95) & 30.00   \\
\cline{1-14}
\multirow{3}{*}{Living room} & Television    & 85.3 & 5.2  & 2.59    & 8.98 (18.36)  & -22.86 (13.31)   & 20.00   & \bb{53.60 (26.06)} & \bb{-8.63 (17.97)}  & \bb{80.00}   & 40.08 (32.14) & -12.22 (28.08) & 50.00   \\
                              & RemoteControl & 69.6 & 4.5  & 1.93    & 9.24 (13.99)  & -13.21 (20.44)   & 30.00   & 18.67 (24.17) & -38.38 (18.29) & 30.00   & \bb{30.14 (28.99)} & \bb{5.81 (25.29)}   & \bb{60.00}   \\
                              & CreditCard    & 42.9 & 4.3  & 1.48    & 3.18 (7.19)   & \textbf{-28.15 (11.70)}   & 10.00   & 0.82 (1.85)   & -53.73 (20.32) & 10.00   & \bb{4.74 (7.19)}   & -35.84 (21.62) & \bb{20.00}   \\
\bottomrule
\end{tabular}
  }
  }
  \caption{
    \textbf{Detection Statistics and Object Search Results Grouped by Target Classes.} Target objects are sorted by average detection range ($r$). We estimated the values for TP, FP, and $r$ by running the vision detector at 30 random camera poses per validation scene. \emph{COS-POMDP} performs more consistently and robustly for hard-to-detect objects, such as \texttt{ScrubBrush}, \texttt{CellPhone}, \texttt{Candle}, and \texttt{Knife}.
  }
  \label{tab:objects}
  \vspace{-0.15in}
\end{table*}

\subsection{Evaluation Metric}
We use three metrics: (1) success weighted by inverse path length (SPL) \cite{anderson2018evaluation}; (2) success rate (SR) and (3) dicounted cumulative rewards (DR).
The SPL for each trial $i$ is defined as $\mbox{SPL}_i=S_i\cdot \ell_i / \max(p_i, \ell_i)$
where $S_i$ is the binary success outcome of the search, $\ell_i$ is the shortest
path between the robot and the target, and $p_i$ is the actual search path. The SPL measures the search performance by taking into account both the success and the efficiency of the search. As a stringent metric, $\ell_i$ uses information about the true object location, but it does not penalize excessive rotations \cite{batra2020objectnav}.
Therefore, we also include the discounted cumulative rewards (DR) metric with $\gamma=0.95$, which takes rotation actions into account.

\subsection{Baselines}
\label{sec:baselines}
Baselines are defined in the caption of Table~\ref{tab:main}.
Note that \emph{Greedy-NBV} is based on \cite{zeng2020semantic} where a weighted particle belief is used to estimate the joint state over all object locations. During planning,
the robot selects the next best viewpoint to navigate towards based on a cost function that considers
both navigation distance and the probability of detecting any object.
This provides a baseline that is in contrast to the sequential decision-making paradigm considered by COS-POMDPs and the modeling of only robot and target states.

\subsection{Implementation Details} 
Objects exist in 3D in AI2-THOR scenes. Since the
robot can tilt its camera within a small range of angles, all methods (except \emph{Random}) estimate target object height among a discrete set of possible height values, \texttt{Above}, \texttt{Below}, and \texttt{Same}, with respect to the camera's current tilt angle.
POMDP-based methods are implemented using the \texttt{pomdp\_py} \cite{pomdp-py-2020} library with the POUCT planner \cite{silver2010monte}. The rollout policy uniformly samples from move actions towards the target or possibly leading to a non-\texttt{null} observation about an object.

\section{Results and Discussions}\label{sec:results}
Our main results by room type are shown in Table~\ref{tab:main}; results over all room types are in the appendix. The performance of \emph{COS-POMDP} is more consistent compared to other baselines at either the best or the second best for all metrics in the four room types.
The performance is broken down by target classes in Table~\ref{tab:objects}. \emph{Greedy-NBV} performs well for \texttt{AlarmClock} in \emph{Bedroom}; it appears to experience less instability in the particle belief as a result of particle reinvigoration. 
\emph{COS-POMDP} appears to be the most robust when the target object has significant uncertainty of being detected correctly, including  \texttt{ScrubBrush}, \texttt{CreditCard}, \texttt{Candle} \texttt{RemoteControl}, \texttt{Knife}, and \texttt{CellPhone}. An example search trial for \texttt{CreditCard} is visualized in Fig.~\ref{fig:example}. For target objects with a true positive detection rate below 40\%, \emph{COS-POMDP} improves the POMDP baseline that ignores correlational information by 42.1\% in terms of the SPL metric ($p=0.028$), and it is more than 2.1 times better than the greedy baseline ($p=0.023$). Both results are statistically significant.  Indeed, when the target object is reliably detectable, such as \texttt{Television}, the ability to detect multiple other objects may actually hurt performance, compared to \emph{Target-POMDP}, due to noise from detecting those other objects and the influence on search behavior.

\textbf{Ablation Studies.}
We also conduct two ablation studies. First, we equip \emph{COS-POMDP} with a groundtruth object detector, as done in \cite{qiu2020learning}, henceforth called \emph{COS-POMDP~(gt)}. This shows the performance when the detections of both the target and correlated objects involve no noise at all. We observe better or competitive performance from using groundtruth detectors across all metrics in all room types. The gain over \emph{COS-POMDP} in terms of SPL is not statistically significant ($p=0.069$).

Additionally, we use correlations estimated using training scenes (\emph{COS-POMDP (est)}) as well as incorrect correlational information that is the reverse of the correct one (\emph{COS-POMDP (wrg)}).  Indeed, using accurate correlations provides the most benefit, while correlations estimated through this naive method could offer benefit compared to using incorrect correlations in some cases (\emph{Bathroom} and \emph{Bedroom}), but can also backfire and hurt performance in others. Therefore, properly learning correlations is important, while leveraging a reliable source of information, for example, from a human at the scene, may offer the most benefit.


\section{Conclusion and Future Work}
In this paper, we formulated the problem of correlational object search as a POMDP (COS-POMDP), and proposed a hierarchical planning algorithm to solve it in practice. Our results show that, particularly for hard-to-detect  objects, our approach offers more robust performance compared to baselines.
Directions for future work include investigating different correlation models, searching in more complex settings that involve \eg, container opening and dynamic objects, and evaluating on a real robot platform.

\noindent\footnotesize{\textbf{Acknowledgements:} This work was supported by awards from Echo Labs, STRAC, and ONR under grant number N00014-21-1-2584. We sincerely thank Leslie Kaelbling and Tomás Lozano-Pérez for their critical and invaluable advice. We also thank Mitchell Wortsman, Yiding Qiu, and Anwesan Pal, and the AI2-THOR developers for helpful clarifications.}


\bibliography{root}

\newpage
\appendix
\normalsize

\subsection{Proof of Theorem 1}
\label{appdx:proof1}
\textbf{Theorem 1.}
    Given any policy $\pi:h_t\rightarrow a$, the distribution of histories is identical between the COS-POMDP and the F-POMDP.

\begin{proof}
  We prove this by induction. When $t=1$, the statement is true because both histories are empty. The inductive hypothesis assumes that the distributions $\Pr(h_t)$ is the same for the two POMDPs at $t\geq 1$. Then, by definition, $\Pr(h_{t+1})=\Pr(h_t,a_t,z_t)=\Pr(z_t|h_t,a_t)\Pr(a_t|h_t)\Pr(h_t)$. Since $\Pr(a_t|h_t)$ is the same under the given $\pi$, we can conclude $\Pr(h_{t+1})$ is identical if the two POMDPs have the same $\Pr(z_t|h_t,a_t)$. We show that this is true as follows.

  First, we sum out the state $s_t$ at time $t$:
  \begin{align}
    &\Pr(z_t|h_t,a_t)= \sum_{s_t}\Pr(s_t,z_t|h_t,a_t)
      \intertext{By definition of conditional probability,}
    &\ \ =\sum_{s_t}\Pr(z_t|s_t,h_t,a_t)\Pr(s_t|h_t,a_t)
      \intertext{Since $s_t$ is does not depend on $a_t$ (which affects $s_{t+1}$),}
    &\ \ =\sum_{s_t}\Pr(z_t|s_t,h_t,a_t)\Pr(s_t|h_t)\label{eq:shared}
  \end{align}
  Suppose we are deriving this distribution for COS-POMDP, denoted as $\PrCOS(z_t|h_t,a_t)$. Then, by definition, the state $s_t=(\xtarget,\srobot)$. Therefore, we can write:
  \begin{align}
      \begin{split}
        &\PrCOS(z_t|h_t,a_t)\\
        &=\sum_{\cosstate}\Pr(z_t|\cosstate,h_t,a_t)\\
        &\quad\quad\quad\quad\quad\quad\times\Pr(\cosstate|h_t)
      \end{split}
      \intertext{Summing out $x_1,\ldots, x_n$,}
      \begin{split}
        &=\sum_{\cosstate}\sum_{x_1,\cdots, x_n}\Pr(x_1,\ldots,x_n,z_t|\cosstate,h_t,a_t)\\
        &\quad\quad\quad\quad\quad\quad\times\Pr(\cosstate|h_t)
      \end{split}
          \intertext{Merging sum,}
      \begin{split}
        &=\sum_{\fullstate}\Pr(x_1,\ldots,x_n,z_t|\cosstate,h_t,a_t)\\
        &\quad\quad\quad\quad\times\Pr(\cosstate|h_t)
      \end{split}
          \intertext{By the definition of conditional probability,}
      \begin{split}
        &=\sum_{\fullstate}\Pr(z_t|x_1,\ldots,x_n,\cosstate,h_t,a_t)\\
        &\quad\quad\quad\quad\times\Pr(x_1,\ldots,x_n|\cosstate,h_t,a_t)\\
        &\quad\quad\quad\quad\times\Pr(\cosstate|h_t)
      \end{split}
          \intertext{Again, because the object locations are independent of $a_t$,}
      \begin{split}
        &=\sum_{\fullstate}\Pr(z_t|x_1,\ldots,x_n,\cosstate,h_t,a_t)\\
        &\quad\quad\quad\quad\times\Pr(x_1,\ldots,x_n|\cosstate,h_t)\\
        &\quad\quad\quad\quad\times\Pr(\cosstate|h_t)
      \end{split}
      \intertext{By the definition of conditional probability again,}
      \begin{split}
        &=\sum_{\fullstate}\Pr(z_t|x_1,\ldots,x_n,\cosstate,h_t,a_t)\\
        &\quad\quad\quad\quad\times\Pr(x_1,\ldots,x_n,\cosstate|h_t)\label{eq:connection}
      \end{split}
    \intertext{Note that $(\xtarget,\srobot,x_1,\ldots,x_n)$ is a state in F-POMDP. Denote the state space of F-POMDP as $\SFPOMDP$. According to Eq (\ref{eq:shared}), we can write the above Eq (\ref{eq:connection}) as}
        &=\sum_{s_t\in\SFPOMDP}\Pr(z_t|s_t,h_t,a_t)\Pr(s_t|h_t)\\
        &=\PrFPOMDP(z_t|h_t,a_t)
  \end{align}
\end{proof}

\subsection{Hierarchical Planning Algorithm}
\label{appendix:hier}
\begin{figure*}[bth]
    \centering
    \includegraphics[width=\linewidth]{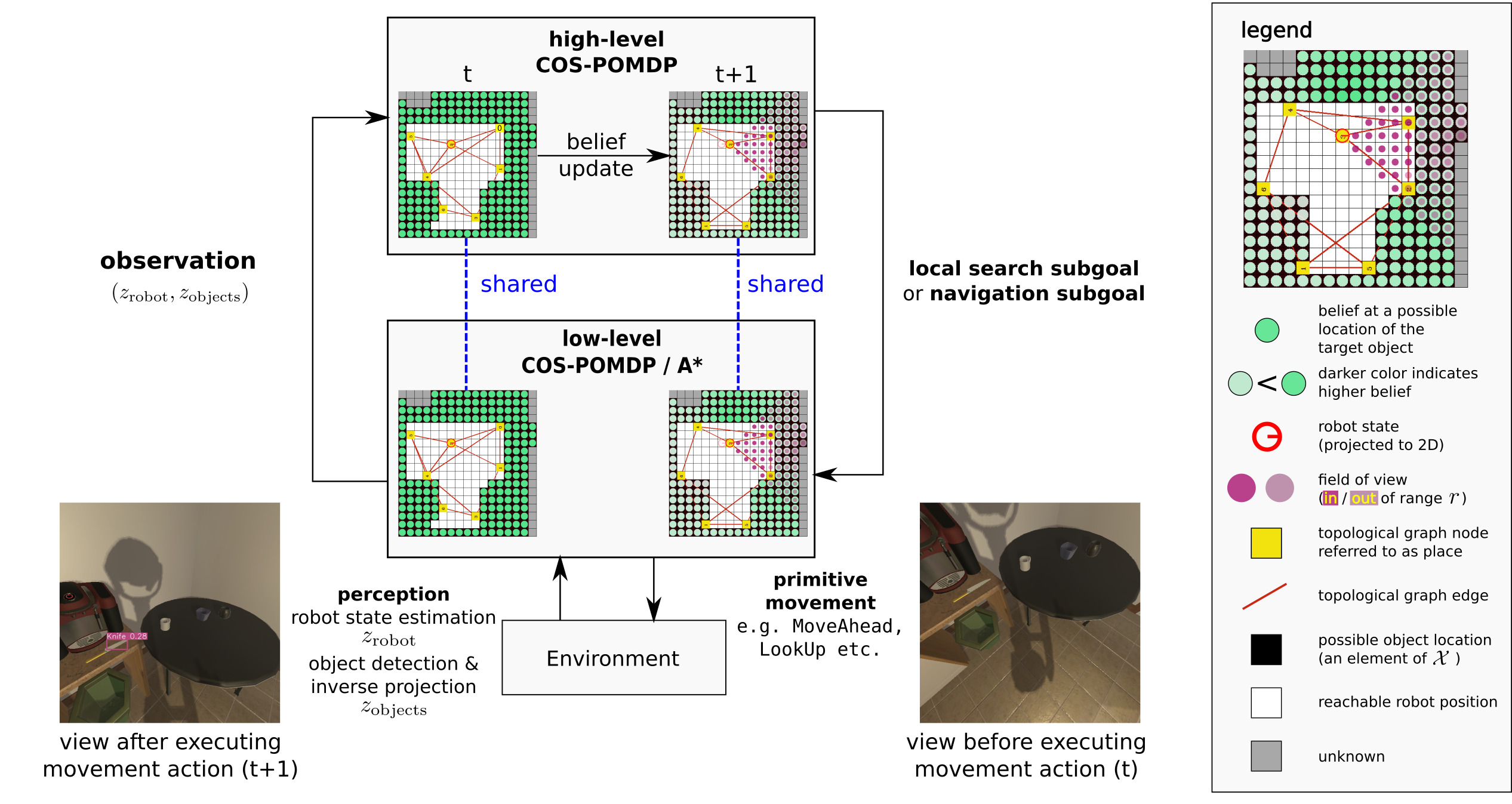}
    \caption{\textbf{Illustration of the Hierarchical Planning Algorithm (with legend).} This is an enlarged version of the figure as Fig.~\ref{fig:method} with a legend. A high-level COS-POMDP plans subgoals that are fed to a low-level planner to produce low-level actions. The belief state is shared across the levels. Both levels plan with updated beliefs at every timestep.
    }
    \label{fig:method-with-legend}
    \vspace{-0.15in}
\end{figure*}

Below, we describe the hierarchical planning algorithm proposed in Sec.~\ref{sec:hier} in detail.
The pseudocode for this algorithm is presented in Algorithm~\ref{alg:hierplanning} and illustrated with legend
in Fig.~\ref{fig:method-with-legend}.

To enable the planning of searching over a large region, we first generate a topological graph, where nodes are places accessible by the robot, and edges indicate navigability between places. 
This is done by the SampleTopoGraph procedure (Algorithm \ref{alg:sampletopo}). In this procedure, the nodes are sampled based on the robot's current belief in the target location $\btarget^t$, and edges are added such that the graph is connected and every node has an out-degree within a given range, which affects the branching factor for planning. An example output is illustrated in Fig.~\ref{fig:method}.

Then, a high-level COS-POMDP $P_{\text{H}}$ is instantiated. The state and observation spaces, the observation model, and the reward model, are as defined in Sec.~\ref{sec:cospomdps}. The move action set and the corresponding transition model are defined according to the generated topological graph. Each move action represents a subgoal of \emph{navigating} to another place, or the subgoal of \emph{searching locally} at the current place.  Both types of subgoals can still be understood as viewpoint-changing actions, except the latter keeps the viewpoint at the same location. For the transition model $T(s',g,s)$ where $g$ represents the subgoal, the resulting viewpoint (\ie, $\srobot'\in s'$) after completing a subgoal is located at the destination of the subgoal with orientation facing the target object location ($\xtarget\in s$). The \texttt{Done} action is also included as a dummy subgoal to match the definition of the COS-POMDP action space (Sec.~\ref{sec:cospomdps}).

At each timestep, a subgoal
is planned using an online POMDP planner, and a low-level planner is instantiated
corresponding to the subgoal. This low-level planner then plans to output an
action $a_t$ from the action set $\mc{A}=\mc{A}_m\cup\{\DECLARE\}$, which is used for
execution.  In our implementation, for \emph{navigation} subgoals, an A$^*$ planner is used, and for \emph{searching locally}, a low-level COS-POMDP $P_{\text{L}}$ is instantiated with the primitive movements $\mc{A}_m$ in its action space. (We use PO-UCT \cite{silver2010monte} as the online POMDP solver in our experiments.)

Upon executing the low-level action $a_t$, the robot receives an observation $z_t\in\mc{Z}$ from its on-board perception modules for robot state estimation and object detection. This observation is used to update the belief of the high-level COS-POMDP, which is shared with the low-level COS-POMDP.

Finally, the process starts over from the first step of sampling a topological graph. If the high-level COS-POMDP plans a new subgoal different the current one, then the low-level planner is re-instantiated.

This algorithm plans actions for execution in an online, closed-loop fashion, allowing reasoning about viewpoint changes both at the level of places in a topological graph as well as fine-grained movements.

\begin{algorithm}[t]
\caption{OnlineHierarchicalPlanning}
\label{alg:hierplanning}
\SetArgSty{textup}
\textbf{Input:} {$P=(\mc{X},\mc{C},\mc{D},\srobotinit, \Srobot, \Orobot, \mc{A}_m,T_m)$}.\\
\textbf{Parameter:} {maximum number of steps $T_{\text{max}}$.}\\
\textbf{Output:} {Action sequence $a_{1},\cdots a_t$} (Sec.~\ref{sec:cos}).\\
$\btarget^1(\xtarget)\gets$ Uniform($\mc{X}$)\;
$\brobot^1(\srobot)\gets\indicator(\srobot=\srobotinit)$\;
$b^1\gets(\btarget^1,\brobot^1)$\;
$t\gets 1$\;
\While{$t \leq T_{\text{max}}$ and $a_{t-1}\neq\texttt{Done}$}{
    $(\mc{V},\mc{E})\gets$ SampleTopoGraph($\mc{X},\Srobot,\btarget^t$)\;
    $P_{\text{H}}\gets$ HighLevelCOSPOMDP($P,\mc{V},\mc{E},b^t$)\;
    subgoal $\gets$ plan POMDP online for $P_{\text{H}}$\;
    \uIf{subgoal is \emph{navigate to a node in $\mc{V}$}}{
        $\srobot\gets \argmax_{\srobot}\brobot(\srobot)$\;
        $a_t \gets$ A$^*$(subgoal, $\srobot$, $\mc{A}_m,T_m$)\;
    }
    \uElseIf{subgoal is \emph{search locally}}{
        $P_{\text{L}}\gets$ LowLevelCOSPOMDP($P$, $b^t$)\;
        $a_t \gets$ plan POMDP online for $P_{\text{L}}$\;
    }
    \ElseIf{subgoal is \emph{Done}}{
        $a_t\gets\texttt{Done}$
    }
    $z_t\gets$ execute $a_t$ and receive observation\;
    $b^{t+1}\gets \text{BeliefUpdate}(b^t,a_t,z_t)$\;
    $t\gets t+1$\;
}
\end{algorithm}

\label{appdx:algs}
\begin{algorithm}[hbt]
\caption{SampleTopoGraph}
\label{alg:sampletopo}
\SetArgSty{textup}
\SetKwInOut{Input}{Input}
\SetKwInOut{Output}{Output}
\SetKwInOut{Parameter}{Parameter}
\textbf{Input:} {$\mc{X},\Srobot,\btarget$}\\
\textbf{Parameter:} {maximum number of nodes $M$, minimum separation between nodes $\dsep$, minimum and maximum out-degrees $\mindeg,\maxdeg$}\\
\textbf{Output:} {A topological graph $(\mc{V},\mc{E})$}\\
\tcp{Obtain mapping from $\srobot$ to a set of closest locations}
\ForEach{$x\in\mc{X}$}{
    $\srobot^{\text{closest}} \gets \argmin_{s_r\in\Srobot}\norm{s_r\text{.pos}-x}$\;
    $U(\srobot^{\text{closest}}) \gets U(\srobot^{\text{closest}}) \cup\{x\}$\;
}
\tcp{Construct probability distribution over $\Srobot$ using $\btarget$}
\ForEach{$\srobot\in\Srobot$}{
    $p(\srobot)\gets\sum_{x\in U(\srobot)}\btarget(x)$
}
\tcp{Construct nodes and edges}
$\mc{V}\gets$ sample $\leq M$ nodes from $\Srobot$ according to $p$ such that any pair of nodes has a minimum distance of $\dsep$\;
$\mc{E}\gets$ add edges between nodes in $\mc{V}$ so that the graph is connected and each node has an out-degree between $\mindeg$ and $\maxdeg$\;
\Return{($\mc{V},\mc{E})$}
\end{algorithm}

Algorithm~\ref{alg:sampletopo} is the pseudocode of the SampleTopoGraph algorithm, implemented for our experiments in AI2-THOR. We set $M=10$, $\dsep=1.0$m, $\mindeg=3$, $\maxdeg=5$. In our implementation, the topological graph is resampled only if the cumulative belief captured by the nodes in the current topological graph, $\sum_{\srobot\in\mc{V}}p(\srobot)$, is below 50$\%$. Otherwise, the same topological graph will be returned.

\begin{table}[htb]
  \centering
    \renewcommand{\arraystretch}{1.2}{
  \resizebox{\linewidth}{!}{%
  \begin{tabular}{l|ccc}
    \specialrule{.8pt}{2pt}{2pt}
    Method          &   SPL (\%)      &  DR              & SR (\%)\\
    \specialrule{.4pt}{2pt}{0pt}
    Random          &  1.59 (2.21)  & -79.63 (3.87)    & 1.59   \\
    Greedy-NBV      &  13.51 (4.45) & -17.80 (4.87)    & 29.37  \\
    Target-POMDP    &  19.81 (5.29) & -20.80 (5.76)    & 38.89  \\
    COS-POMDP       &  \textbf{24.53 (5.94)} & \textbf{-13.44 (5.64)}    & \textbf{43.65}  \\
    \hline
    COS-POMDP (gt)  &  \textbf{30.57 (6.24)} & \textbf{-11.44 (6.16)}    & \textbf{52.38}  \\
    \hline
    COS-POMDP (est) &  14.23 (4.78) & -32.08 (5.59)    & 29.36  \\
    COS-POMDP (wrg) &  16.76 (5.18) & -19.86 (4.64)    & 31.74  \\
    \bottomrule
  \end{tabular}
}}
  \caption{ \textbf{Results over all room types.} Table formatting follows Table~\ref{tab:main}. See text in Appendix \ref{appendix:results} for discussion. }
  \label{tab:overall}
  \vspace{-0.2in}
\end{table}

\subsection{Additional Results and Discussions}
\label{appendix:results}
The performance over all scenes and target classes are shown in Table~\ref{tab:overall}.
In summary, \emph{COS-POMDP} outperforms the baselines across all three metrics. Based on Table~\ref{tab:objects}, we also observe an advantage for \emph{COS-POMDP} for objects that are detected at a closer distance on average ($r$). In particular, The performance gain over \emph{Greedy-NBV} is statistically significant ($p<0.001$) in terms of SPL, and the performance gain over \emph{Target-POMDP} is statistically significant ($p=0.04$) in terms of discounted cumulative rewards. In addition, \emph{COS-POMDP} is significantly better than \emph{COS-POMDP (est)} ($p=0.002$) and \emph{COS-POMDP (wrg)} ($p=0.012$) in terms of SPL. \emph{COS-POMDP (gt)} outperforms \emph{COS-POMDP} in SPL but is not significant ($p=0.069$). \emph{COS-POMDP (est)} performs worse than wrong \emph{COS-POMDP (wrg)} in \emph{Kitchen} and \emph{Living room}. Our observation is that scenes in those room types have greater variance in size and layout, making estimated correlations less desirable in validation scenes, and they may contain multiple instances of some object classes such that search by \emph{COS-POMDP (wrg)} may actually benefit from belief update using the reverse of correct correlational information since it may in fact increase the probability over one of the true target locations.

\begin{table}[hbt]
  \setlength{\tabcolsep}{4pt}
  \centering
\renewcommand{\arraystretch}{1.2}{
  \resizebox{0.85\linewidth}{!}{%
  \begin{tabular}{ll|ccc}
\specialrule{.8pt}{2pt}{0pt}
        Room Type             & Correlated Object Class  & TP   & FP   & $r$ (m) \\
\specialrule{.4pt}{0pt}{0pt}
    \multirow{5}{*}{Bathroom} & Mirror                   & 76.9 & 3.7  & 2.10\\
                              & ToiletPaperHanger        & 84.4 & 1.5  & 1.96\\
                              & Towel                    & 79.4 & 2.7  & 1.88\\
                              & Toilet                   & 86.3 & 3.5  & 1.81\\
                              & SoapBar                  & 73.2 & 1.8  & 1.53\\
\cline{1-5}
    \multirow{5}{*}{Bedroom}  & DeskLamp                 & 89.5 & 2.6  & 2.41\\
                              & Bed                      & 63.5 & 0.6  & 2.39\\
                              & Mirror                   & 86.0 & 0.6  & 2.27\\
                              & LightSwitch              & 76.3 & 2.8  & 2.26\\
                              & Laptop                   & 75.9 & 1.2  & 2.19\\
\cline{1-5}
    \multirow{5}{*}{Kitchen}  & LightSwitch              & 90.0 & 3.9  & 2.57\\
                              & Microwave                & 75.3 & 5.6  & 2.31\\
                              & StoveKnob                & 82.8 & 5.6  & 2.00\\
                              & Lettuce                  & 98.6 & 0.3  & 1.98\\
                              & Plate                    & 60.6 & 3.2  & 1.90\\

\cline{1-5}
    \multirow{6}{*}{Living room}  & FloorLamp            & 71.7 & 5.1 & 3.44\\
                              & Painting                 & 85.2 & 4.0 & 3.18\\
                              & LightSwitch              & 80.6 & 1.5 & 3.10\\
                              & HousePlant               & 82.9 & 3.9 & 3.00\\
                              & Pillow                   & 67.4 & 2.8 & 2.84\\
                              & Laptop                   & 66.3 & 2.6 & 2.24\\
\specialrule{.8pt}{0pt}{0pt}
\end{tabular}
  }
  }
  \caption{ \textbf{Detection Statistics for Correlated Object Classes.} TP: true positive rate (\%); FP: false positive rate (\%); $r$: average distance to the true positive detections (m). We estimated these values by running the vision detector at 30 random camera poses per validation scene. The correlated object classes for each room type are sorted by average detection range.
  }
  \label{tab:detcorr}
\end{table}

\subsection{Detection Statistics for Correlated Object Classes}
\label{appdx:detcorr}
The correlated object classes are chosen to have, in the validation scenes, at
least 60\% true positive rate and generally above 70\%, and around or below 5\%
false positive rate and an average true positive detection distance of around 2m or
more. This contrasts the target classes where either the true positive rate is
below 60\% (with many below 50\%), false
positive rate around 5-10\%, or the average true positive detection distance of
around or less than 2.5m,

Table~\ref{tab:detcorr} shows the detection statistics for correlated object classes. The detection statistics of target object classes can be found in Table~\ref{tab:objects}.

\end{document}